\documentclass[conference]{IEEEtran}
\IEEEoverridecommandlockouts

\usepackage{cite}
\usepackage{amsmath,amssymb,amsfonts}
\usepackage{graphicx}
\usepackage{booktabs}
\usepackage[table]{xcolor}
\usepackage{array}
\usepackage{hyperref}
\usepackage{algorithm}
\usepackage{algorithmic}
\usepackage{comment}
\usepackage[margin=1in]{geometry}
\usepackage{tikz}
\usetikzlibrary{calc, bayesnet, positioning}
\usepackage{xcolor}
\def\BibTeX{{\rm B\kern-.05em{\sc i\kern-.025em b}\kern-.08em
    T\kern-.1667em\lower.7ex\hbox{E}\kern-.125emX}}

\begin{document}

\title{Variational Self-Supervised Learning}

\author{
\IEEEauthorblockN{Mehmet Can Yavuz}
\IEEEauthorblockA{\textit{Faculty of Engineering and Natural Sciences} \\
\textit{Işık University}\\
İstanbul, Türkiye \\
mehmetcan.yavuz@isikun.edu.tr}
\and
\IEEEauthorblockN{Berrin Yanıkoğlu}
\IEEEauthorblockA{\textit{Faculty of Engineering and Natural Sciences} \\
\textit{Sabancı University}\\
İstanbul, Türkiye \\
berrin@sabanciuniv.edu}
}

\maketitle

\begin{abstract}
We present Variational Self-Supervised Learning (VSSL), a novel framework that combines variational inference with self-supervised learning to enable efficient, decoder-free representation learning. Unlike traditional VAEs that rely on input reconstruction via a decoder, VSSL symmetrically couples two encoders with Gaussian outputs. A momentum-updated teacher network defines a dynamic, data-dependent prior, while the student encoder produces an approximate posterior from augmented views. The reconstruction term in the ELBO is replaced with a cross-view denoising objective, while preserving the analytical tractability of Gaussian KL divergence terms. We further introduce cosine-based formulations of KL and log-likelihood terms to enhance semantic alignment in high-dimensional latent spaces. Experiments on CIFAR-10, CIFAR-100, and ImageNet-100 show that VSSL achieves competitive or superior performance to leading self-supervised methods, including BYOL and MoCo V3. VSSL offers a scalable, probabilistically grounded approach to learning transferable representations without generative reconstruction, bridging the gap between variational modeling and modern self-supervised techniques.
\end{abstract}

\begin{IEEEkeywords}
self-supervised learning, variational inference, representation learning, encoder-only models
\end{IEEEkeywords}

\section{Introduction}
Variational inference has become a foundational paradigm in machine learning, enabling scalable approximations of posterior distributions in latent variable models. The variational autoencoder (VAE), introduced by Kingma and Welling (2013), exemplifies this approach by optimizing the Evidence Lower Bound (ELBO), which balances a reconstruction term—computed via a decoder—with a Kullback-Leibler (KL) divergence regularizer aligning the approximate posterior to a prior \cite{kingma2013auto}. While effective, the decoder’s role in reconstructing input data introduces computational complexity and assumes reconstruction as a prerequisite for meaningful representations, a constraint that may not always be necessary. In parallel, self-supervised learning has gained prominence for its ability to extract high-quality features without explicit reconstruction, often outperforming supervised methods in representation learning tasks. For instance, contrastive methods like SimCLR \cite{chen2020simple} leverage mutual information between augmented views of data to learn robust features, while predictive frameworks such as BYOL \cite{grill2020bootstrap} achieve similar success by aligning representations across networks without negative samples. It is well-established that self-supervised methods excel at learning quality features, capturing intricate patterns and transferable representations more effectively than traditional supervised or generative approaches \cite{jing2020self}.

Motivated by these insights, we propose a novel variational self-supervised learning framework that eschews the decoder entirely, instead symmetrically coupling two encoders with Gaussian-distributed outputs. In our approach, each encoder generates an approximate posterior conditioned on the input and predicts the prior of the other’s latent representation, derived from a momentum-based network. This formulation retains the analytical tractability of Gaussian KL divergence \cite{blei2017variational}, while replacing the reconstruction objective with a cross-predictive task. By integrating variational principles with self-supervised symmetry and momentum encoder, our method offers a decoder-free alternative that aligns with the strengths of recent representation learning paradigms.

\begin{figure}[ht]
\centering
\resizebox{\linewidth}{!}{%
\begin{tikzpicture}[node distance=1.8cm and 2.4cm]
  \node[obs] (x1) {$x_1$};
  \node[obs, right=4.2cm of x1] (x2) {$x_2$};
  \node[latent, below=1.8cm of $(x1)!0.5!(x2)$] (z) {$z$};
  \node[latent, below left=0.6cm and 2.8cm of x2] (zhat) {$\hat{z}$};
  \node[latent, below right=0.6cm and 2.8cm of x1] (phi) {$\phi$};
  \node[latent, below=2.2cm of x1] (thetat) {$\theta_t$};
  \node[latent, below=2.2cm of x2] (thetas) {$\theta_s$};

  \edge{x1}{thetat};
  \edge{thetas}{phi}
  \edge{thetat}{z};
  \edge{x2}{thetas};
  \edge{thetas}{z};

  \draw (z) to[bend left] (zhat);   
  \draw (z) to[bend right] (zhat);  
  
  \edge{phi}{zhat};

  \draw[dashed,->] (thetas) -- node[above left=-0.5cm and -1.6cm] {\scriptsize Exponential Moving Average} (thetat);

  \node[below=0.05cm of thetat] {\scriptsize $q_{\theta_t}(z \mid x_1)$};
  \node[below=0.05cm of thetas] {\scriptsize $q_{\theta_s}(z \mid x_2)$};
  \node[above=0.05cm of zhat] {\scriptsize denoiser $p_{\phi}(z \mid \hat{z})$};
\end{tikzpicture}
}
\caption{Directed graphical model for the VSSL framework. Observations \(x_1\) and \(x_2\) are encoded via parameterized inference networks \(\theta_t\) and \(\theta_s\), producing the latent representation \(z\). The student path \(q_{\theta_s}(z \mid x_2)\) updates the teacher path \(q_{\theta_t}(z \mid x_1)\) through exponential moving average (EMA). A denoising network \(p_\phi(\hat{z} \mid z)\) refines the latent, producing a denoised representation \(\hat{z}\) for self-supervised learning objectives.}
\end{figure}
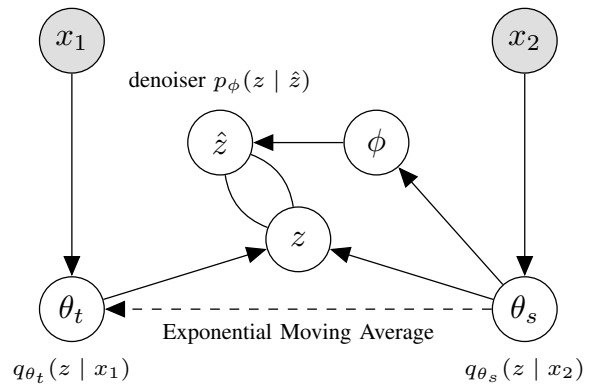

\section{Related Work}

Variational inference has emerged as a cornerstone of probabilistic modeling in machine learning, offering a scalable framework for approximating posterior distributions in latent variable models. The variational autoencoder (VAE), introduced by Kingma and Welling \cite{kingma2013auto}, serves as a seminal example, optimizing the Evidence Lower Bound (ELBO) by balancing a reconstruction term with a Kullback-Leibler (KL) divergence regularizer. The VAE's reliance on a decoder to reconstruct input data, while effective for generative tasks, introduces computational overhead and assumes reconstruction is essential for learning meaningful representations—a constraint that may not always align with representation learning objectives. Subsequent advancements, such as $\beta$-VAE \cite{higgins2017beta} and InfoVAE \cite{zhao2019infovae}, have refined this framework by adjusting the trade-off between reconstruction fidelity and latent regularization, yet they largely retain the decoder-centric structure.

In parallel, self-supervised learning has gained traction as a powerful paradigm for learning high-quality representations without explicit supervision or reconstruction. Contrastive methods, like SimCLR \cite{chen2020simple}, leverage mutual information maximization between augmented views of the same input to learn robust features, often outperforming supervised approaches in downstream tasks like image classification and transfer learning \cite{jing2020self}. However, these methods typically require negative samples, which increases computational complexity. Predictive self-supervised frameworks, such as BYOL \cite{grill2020bootstrap} and SimSiam \cite{chen2021exploring}, eliminate the need for negative samples by aligning representations across two networks—one often updated via momentum—demonstrating that reconstruction-free objectives can yield competitive representations. These approaches highlight the potential of encoder-only architectures, though they generally lack the probabilistic grounding provided by variational methods.

Efforts to bridge variational inference and self-supervised learning have been explored in prior work. For instance, Variational Predictive Coding \cite{buckley2017free} integrates predictive objectives into a variational framework, but it still typically assumes a generative component akin to a decoder. Similarly, Contrastive Variational Autoencoders \cite{wang2022contrastvae} merge these concepts, yet often retain the decoder for reconstruction or related generative tasks. More recently, works like Barlow Twins \cite{zbontar2021barlow} and VICReg \cite{bardes2022vicreg} have proposed non-contrastive objectives based on redundancy reduction and variance preservation, aligning representations without reconstruction, though they do not explicitly adopt a variational formulation. Other self-supervised variational learning approaches \cite{yavuz2024self, yavuz2022vcl} combines contrastive learning with variational objectives. These methods underscore the shift toward decoder-free representation learning but often sacrifice the analytical tractability and probabilistic interpretability associated with variational inference.

\begin{figure*}[thb]
  \begin{center}
    \includegraphics[width=\textwidth]{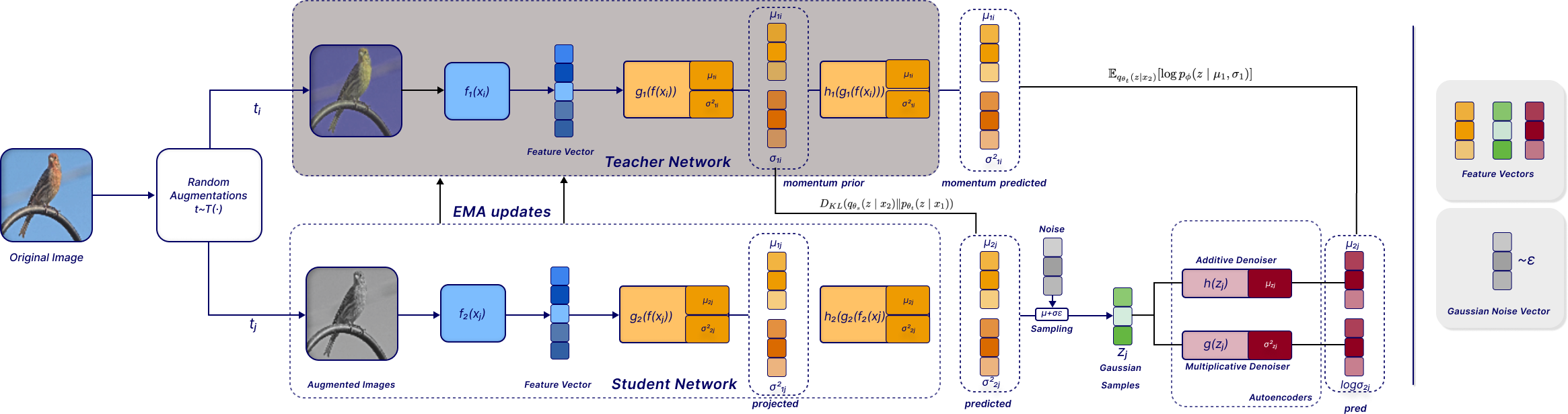}
  \end{center}
  \caption{Overview of the variational self-supervised learning framework for unsupervised representation learning using variational objectives. An original image is augmented into two views, $t_i$ and $t_j$, processed by a momentum-updated teacher network and a student network, respectively. Both networks encode the image into feature vectors and variational distributions parameterized by $\mu$ and $\log\sigma$. The teacher outputs serve as a prior for the student’s posterior via a KL divergence minimization. Gaussian sampling from the student posterior allows further processing through an autoencoder, enforcing consistency and regularization in the learned latent space.}
  \label{Fig:VariationalContrastive}
\end{figure*}

Our proposed framework builds on these insights by fully eschewing the decoder and symmetrically coupling two encoders with Gaussian-distributed outputs, inspired by the momentum-based priors of BYOL \cite{grill2020bootstrap}. Unlike traditional VAEs, we replace the reconstruction term with a cross-predictive task where each encoder’s approximate posterior predicts the prior of the other, leveraging the closed-form KL divergence for Gaussian distributions \cite{blei2017variational}. This approach aims to marry the probabilistic rigor of variational inference with the feature-learning strengths of self-supervised methods, distinguishing it from prior hybrid models that retain generative components or lack a symmetric encoder design. By eliminating the decoder while preserving analytical tractability, our method offers a potentially novel and efficient alternative for latent representation learning.

\section{Methodology}
In this section, we detail the methodology for a novel variational self-supervised learning (VSSL) framework that eliminates the need for a decoder. Instead, it symmetrically couples two encoders with Gaussian-distributed outputs to learn meaningful latent representations. By integrating variational inference with self-supervised learning principles, this approach replaces the traditional reconstruction objective with a cross-predictive task, leveraging the analytical tractability of Gaussian distributions.

\subsection{Variational Formulation}

The variational autoencoder (VAE) is a generative model designed to learn a probabilistic latent representation of data. For a given data point \( x \), the VAE optimizes the Evidence Lower Bound (ELBO) on the marginal log-likelihood \( \log p(x) \):

\begin{equation}
\mathcal{L}(x) = \mathbb{E}_{q(z \mid x)} [\log p(x \mid z)] - D_{\mathrm{KL}}(q(z \mid x) \| p(z))
\end{equation}

Here:
\begin{itemize}
    \item \( q(z \mid x) \): approximate posterior distribution, parameterized by an encoder network.
    \item \( p(x \mid z) \): likelihood, parameterized by a decoder network.
    \item \( p(z) \): prior over the latent variable \( z \), typically a standard normal \( \mathcal{N}(0, I) \).
\end{itemize}

The ELBO consists of two terms:
\begin{enumerate}
    \item \textbf{Reconstruction Term}: \( \mathbb{E}_{q(z \mid x)} [\log p(x \mid z)] \), encouraging accurate reconstruction of \( x \) from \( z \).
    \item \textbf{KL Divergence Term}: \( D_{\mathrm{KL}}(q(z \mid x) \| p(z)) \), regularizing the posterior to align with the prior.
\end{enumerate}

\subsection{Self-Supervised Learning Context}

Self-supervised learning leverages supervisory signals from the data itself, often through multiple views or augmentations. In this work, we adapt the VAE framework to self-supervised learning using a teacher-student mechanism and a denoising task. Two views, \( x_1 \) and \( x_2 \) (e.g., augmentations of the same sample), are used to improve the robustness of the latent representations.

\subsection{EMA-Updated Teacher Network as Prior}

In standard VAEs, the prior \( p(z) \) is fixed as \( \mathcal{N}(0, 1) \). We introduce a dynamic, data-dependent prior using a teacher network with parameters \( \theta_t \), updated via an exponential moving average (EMA) of the student encoder parameters \( \theta_s \):

\begin{equation}
\theta_t \leftarrow \tau \theta_t + (1 - \tau) \theta_s
\end{equation}

where \( \tau \in [0, 1) \) is a decay factor. The teacher processes view \( x_1 \) to define the prior:

\begin{equation}
p_{\theta_t}(z \mid x_1) = \mathcal{N}(z \mid \mu_t(x_1), \sigma_t^2(x_1))
\end{equation}

The KL divergence in the ELBO becomes:

\begin{equation}
D_{\mathrm{KL}}(q_{\theta_s}(z \mid x_2) \| p_{\theta_t}(z \mid x_1))
\end{equation}

where \( q_{\theta_s}(z \mid x_2) = \mathcal{N}(z \mid \mu_s(x_2), \sigma_s^2(x_2)) \) is the student posterior. This term encourages consistency between the student’s encoding of \( x_2 \) and the teacher’s prior from \( x_1 \), stabilized by the EMA update.

\subsection{Reconstruction Term with Denoiser Networks}

We replace the standard reconstruction term with a denoising objective. Denoiser networks, parameterized by \( \phi \), predict distributional parameters from view \( x_1 \):

\begin{equation}
p_{\phi}(z \mid \mu_1, \sigma_1) = \mathcal{N}(z \mid \mu_{\phi}(x_1), \sigma_{\phi}^2(x_1))
\end{equation}

Letting \( \mu_1 = \mu_{\phi}(x_1) \), \( \sigma_1 = \sigma_{\phi}(x_1) \), the new reconstruction objective becomes:

\begin{align}
&\mathbb{E}_{q_{\theta_t}(z \mid x_2)} [\log p_{\phi}(z \mid \mu_1, \sigma_1)]
\end{align}

Here, \( z \) is sampled using the reparameterization trick: \( z = \mu + \sigma \cdot \epsilon \), with \( \epsilon \sim \mathcal{N}(0, 1) \).

This jointly trains:
\begin{itemize}
    \item The teacher encoder, \( q_{\theta_t}(z \mid x_2) \), to map \( x_2 \) to latent space.
    \item The denoiser networks, \( p_{\phi}(z \mid \mu_1, \sigma_1) \), to infer latent likelihood from \( x_1 \).
\end{itemize}

This setup enforces probabilistic alignment between the latent codes of \( x_1 \) and \( x_2 \), avoiding direct input reconstruction.

\subsection{Derivation of the Self-Supervised ELBO}

We derive the modified ELBO for this self-supervised framework. Given views \( x_1 \) and \( x_2 \), we maximize the likelihood \( \log p_{\phi, z \sim q_{\theta_t}(z \mid x_2)}(z \mid \mu_1, \sigma_1) \). Applying Jensen’s inequality gives:

\begin{align}
&\log p_{\phi, z \sim q_{\theta_t}(z \mid x_2)}(z \mid \mu_1, \sigma_1) \geq \nonumber\\
&\qquad + \mathbb{E}_{z \sim q_{\theta_t}(z \mid x_2)} [\log p_{\phi}(z \mid \mu_1, \sigma_1)] \nonumber\\
&\qquad - D_{\mathrm{KL}}(q_{\theta_s}(z \mid x_2) \| p_{\theta_t}(z \mid x_1)) 
\end{align}

\subsection{Overall Objective}

Putting it all together, our variational self-supervised loss is
\begin{align}
\mathcal{L}_{\mathrm{VSSL}}
&=
\sum_{v_1=1}^2
\sum_{\substack{v_2=1}}^2
\Bigl[
D_{\mathrm{KL}}\bigl(q_{\theta_s}(z\mid x_{v_1}) \,\|\,
p_{\theta_t}(z\mid x_{v_2})\bigr)
\nonumber\\[-0.5ex]
&\quad
-\;\mathbb{E}_{z\sim q_{\theta_s}(z\mid x_{v_1})}
\bigl[\log p_{\phi}(z\mid \mu_{v_2},\sigma_{v_2})\bigr]
\Bigr],
\label{eq:vssl_loss}
\end{align}
and we solve
\[
\min_{\theta_s,\phi}\;\mathcal{L}_{\mathrm{VSSL}}.
\]

\subsection{Implementation Details}

\textbf{Augmentation.} VSSL \footnote{\href{https://github.com/convergedmachine/vssl-solo-learn}{GitHub Repository: github.com/convergedmachine/vssl-solo-learn}} employs SimCLR-style augmentations \cite{chen2020simple}, including random cropping to 224×224, horizontal flips, color jitter (brightness, contrast, saturation, hue), and optional grayscale conversion to enhance view diversity.

\textbf{Architecture.} The framework includes an online student network and a momentum-updated teacher network. \textit{Projectors, Predictors, and Denoisers:} Each component consists of a two-layer MLP with batch normalization and ReLU activations, projecting input features into mean and variance parameters. A primary predictor further refines these representations, while two denoising networks independently predict mean and variance for the latent reconstruction objective.

\textbf{Variational Modeling.} Latent variables are modeled as Gaussian distributions with unconstrained mean and variance parameters and an approximation to actual re-parametrization trick:

\[
\epsilon \sim \left| \mathcal{N}(0, 1) \right|, \quad z = \mu + \sigma^2 \cdot \epsilon
\]

\textbf{Training.} Each training step involves:
\begin{enumerate}
    \item Computing the student posterior and teacher prior.
    \item Estimating the denoising likelihood from the latent features.
    \item Calculating the KL divergence between the student and teacher distributions.
\end{enumerate}
Gradients are used to update student parameters, while teacher parameters are updated using exponential moving average (EMA) of the student weights.

\textbf{Cosine-Based KL Divergence and Log-Likelihood.} To enhance semantic alignment in high-dimensional latent spaces, we propose cosine-based alternatives to classical KL divergence and log-likelihood formulations, leveraging the angular properties of cosine similarity.

Let \( S_{\beta}(a, b) = \mathrm{softplus}_{\beta}(-\cos(a, b)) \), where \( \cos(a, b) \) denotes the cosine similarity between vectors \( a \) and \( b \), and \( \mathrm{softplus}_{\beta} \) is the softplus function scaled by the parameter \( \beta \).

\textit{Cosine-Based KL Divergence:}
\noindent This formulation defines a divergence measure that replaces traditional Euclidean distance terms with cosine-based similarities, promoting angular alignment in latent distributions. The softplus-transformed cosine terms ensure smooth optimization, with a sharper focus on directional consistency between the mean and variance vectors.

\begin{align}
\tilde{D}_{\mathrm{KL}}^{\cos}(\mu_1, \mu_2, \sigma_1^2, \sigma_2^2) 
= \frac{1}{2} \Big[
\log \big(S_{3}(\sigma_1^2, \sigma_2^2)\big) \nonumber\\
+ \big(S_{3}(\mu_1, \mu_2)\big)^2 \nonumber\\
+ S_{3}(\sigma_1^2, \sigma_2^2) - 1 \Big]
\end{align}

\textit{Expected Cosine-Based Log-Likelihood:}
\noindent The expected log-likelihood is reformulated to leverage cosine similarities, integrating both angular alignment of mean vectors and the spread (variance) of latent representations. This approach emphasizes semantic coherence between distributions, providing a robust alternative to standard log-likelihood terms in high-dimensional embedding spaces.

\begin{align}
\tilde{\mathcal{L}}^{\cos}_{\mathrm{ll}}(\mu_1, \mu_2, \sigma_1^2, \sigma_2^2) 
= \log\big(S_{1}(\sigma_1^2, \sigma_2^2)\big) \nonumber\\
+ 4\, S_{1}(\sigma_1^2, \sigma_2^2) \nonumber\\
+ \big(S_{1}(\mu_1, \mu_2)\big)^2 \cdot S_{1}(\sigma_1^2, \sigma_2^2)
\end{align}

\bigskip

\noindent \textit{Cosine similarity}, as employed here, offers a resilient angular metric that remains effective in high-dimensional latent spaces. By substituting Euclidean-based alignment with cosine-driven measures, this framework enhances representational alignment and stability in variational inference models.

\begin{algorithm}
\caption{Variational Self-Supervised Learning}
\begin{algorithmic}[1]
\REQUIRE Dataset $\mathcal{D} = \{x^{(i)}\}_{i=1}^{N}$
\ENSURE Trained student and teacher parameters $\phi, \theta$
\STATE Initialize student $\phi$ and teacher $\theta$ with momentum copies
\FOR{each epoch}
    \FOR{each mini-batch $\{x_1, x_2, \dots, x_k\}$ from $\mathcal{D}$}
        \FOR{each view $x_i$}
            \STATE Encode using student: $\mu_{i}, \sigma_{i} = \text{Pj}_\phi(f(x_i))$
            \STATE Predict posterior: $\hat{\mu}_i, \hat{\sigma}_i = \text{Pd}_\phi(\mu_i, \sigma_i)$
        \ENDFOR
        \FOR{each view $x_j$}
            \STATE Encode using teacher: 
            \STATE $\mu_{j}^\text{mom}, \sigma_{j}^\text{mom} = \text{Pj}_\theta(f(x_j))$
            \STATE Predict momentum prior: 
            \STATE $\hat{\mu}_j^\text{mom}, \hat{\sigma}_j^\text{mom} = \text{Pd}_\theta(\mu_j^\text{mom}, \sigma_j^\text{mom})$
        \ENDFOR
        \STATE Compute KL divergence between posteriors and priors: $\mathcal{L}_{\text{KL}}$
        \STATE Denoise student samples to get $(\mu^\text{recon}, \sigma^\text{recon})$
        \STATE Compute expected log-likelihood loss: $\mathcal{L}_{\text{ll}}$
        \STATE Compute total loss: $\mathcal{L} = \mathcal{L}_{\text{KL}} - \mathcal{L}_{\text{ll}}$
        \STATE Update $\phi$ using gradients from $\nabla_{\phi} \mathcal{L}$
        \STATE Update $\theta$ using EMA of $\phi$
    \ENDFOR
\ENDFOR
\end{algorithmic}
\label{alg:vssl}
\end{algorithm}

Algorithm~\ref{alg:vssl} outlines the training procedure for the proposed Variational Self-Supervised Learning (VSSL) framework. It iteratively updates the student and teacher networks by aligning their latent posterior distributions through KL divergence while optimizing an expected log-likelihood objective to enhance representation quality.

\begin{table*}[htb!]
\centering
\small 
\begin{tabular}{@{\extracolsep{\fill}}lcccc@{}} 
\toprule
\textbf{Method} & \textbf{CIFAR-10 Acc@1} & \textbf{CIFAR-100 Acc@1} & \textbf{ImageNet-100 (Online)} & \textbf{ImageNet-100 (Offline)} \\
\midrule
\rowcolor[gray]{0.95} Barlow Twins \cite{zbontar2021barlow} & 92.10 & \textbf{70.90}  & \underline{80.38}  & 80.16  \\
BYOL \cite{grill2020bootstrap} & \underline{92.58}  & \underline{70.46}  & 80.16 & \underline{80.32}  \\
\rowcolor[gray]{0.95} DeepCluster V2 \cite{caron2021unsupervised} & 88.85 & 63.61 & 75.36 & 75.40 \\
DINO \cite{oquab2023dinov2} & 89.52 & 66.76 & 74.84 & 74.92 \\
\rowcolor[gray]{0.95} MoCo V2+ \cite{chen2020improved} & 92.94 & 69.89 & 78.20 & 79.28 \\
MoCo V3 \cite{chen2021mocov3}& \underline{93.10}  & 68.83 & \underline{80.36}  & \underline{80.36}  \\
\rowcolor[gray]{0.95} NNCLR \cite{dwibedi2021little} & 91.88 & 69.62 & 79.80 & 80.16   \\
ReSSL \cite{zheng2021ressl} & 90.63 & 65.92 & 76.92 & 78.48 \\
\rowcolor[gray]{0.95} SimCLR \cite{chen2020simple}& 90.74 & 65.78 & 77.04 & - \\
SimSiam \cite{chen2021exploring} & 90.51 & 66.04 & 74.54 & 78.72 \\
\rowcolor[gray]{0.95} SwAV \cite{caron2020unsupervised} & 89.17 & 64.88 & 74.04 & 74.28 \\
VICReg \cite{bardes2022vicreg} & 92.07 & 68.54 & 79.22 & 79.40 \\ \midrule
\rowcolor[gray]{0.95} VSSL (this work) & \textbf{93.35}  & \textbf{70.75}  & \textbf{81.15}  & \textbf{81.01}  \\ 	
\bottomrule
\end{tabular}
\caption{Top-1 classification accuracy (\%) of various self-supervised learning methods on CIFAR-10, CIFAR-100, and ImageNet-100 under both online and offline linear evaluation protocols. The best results are shown in \textbf{bold}, while the second- and third-best results are \underline{underlined}. Our proposed method, VSSL, achieves competitive or state-of-the-art performance across multiple datasets and settings.}
\label{tab:ssl_comparison}
\end{table*}

\section{Experimental Evaluation}

We evaluate the proposed VSSL method using the \texttt{solo-learn} framework~\cite{sololearn} across three benchmark datasets: CIFAR-10, CIFAR-100~\cite{krizhevsky2009cifar}, and ImageNet-100 ~\cite{deng2009imagenet}. Model performance is assessed using both \emph{online} and \emph{offline} (only for ImageNet-100) linear evaluation protocols, reporting top-1 classification accuracy.

To ensure a fair and meaningful comparison, we conduct extensive hyperparameter tuning for VSSL. For the comparison table we use high-quality open-source baselines results by \texttt{solo-learn}. In many cases, our reproduced baselines by solo-learn achieve results that are stronger than those originally reported, enabling a more robust evaluation setting.

The full results are summarized in Table~\ref{tab:ssl_comparison}, where we highlight the best, second-best, and third-best performances across all datasets and settings.

\section{Results and Discussion}

Table~\ref{tab:ssl_comparison} presents the top-1 classification accuracy (\%) for a selection of prominent self-supervised learning (SSL) methods across three benchmark datasets: CIFAR-10, CIFAR-100, and ImageNet-100. Evaluations are conducted under both \textit{online} and \textit{offline} linear evaluation protocols to ensure a comprehensive assessment of learned representations.

For CIFAR-10, the majority of methods achieve over 90\% accuracy, demonstrating the maturity of SSL techniques on this relatively simple dataset. Notably, our proposed method, VSSL, achieves the highest top-1 accuracy at \textbf{93.35\%}, outperforming the strong baseline MoCo V3 (\underline{93.10\%}) and BYOL (\underline{92.58\%}). 

On the more challenging CIFAR-100 dataset, which contains a higher number of classes and greater inter-class variability, Barlow Twins achieves the highest accuracy of \textbf{70.90\%}. VSSL closely follows with \textbf{70.75\%}, marking it as highly competitive and outperforming other methods such as MoCo V2+ (69.89\%) and VICReg (68.54\%). This demonstrates the effectiveness of VSSL in handling fine-grained classification tasks.

For the ImageNet-100 dataset, which provides a larger-scale evaluation, results are reported under both online and offline linear probing protocols. VSSL consistently outperforms all competing methods, achieving top-1 accuracies of \textbf{81.15\%} (online) and \textbf{81.01\%} (offline). These results surpass the second-best method, MoCo V3, which achieves \underline{80.36\%} in both settings. The strong performance of VSSL across both evaluation modes indicates the robustness and generalizability of its learned representations.

Overall, the results suggest that VSSL not only achieves state-of-the-art performance across multiple datasets but also demonstrates superior generalization across diverse evaluation protocols. Its competitive accuracy on CIFAR-100 and ImageNet-100, which feature more complex classification scenarios, highlights the efficacy of variational self-supervised learning strategies. The consistent outperformance of established methods such as MoCo V3, BYOL, and Barlow Twins further underscores the potential of VSSL as a versatile and scalable SSL framework.

\section{Conclusion}

We presented VSSL, a novel variational self-supervised learning framework that eliminates the need for a decoder by adopting a symmetric encoder architecture and aligning representations through Gaussian latent distributions. By integrating variational inference with momentum-based contrastive learning, VSSL replaces traditional reconstruction-based objectives with a denoising cross-predictive task, maintaining analytical tractability while enabling efficient and scalable representation learning.

Empirical results across CIFAR-10, CIFAR-100, and ImageNet-100 demonstrate that VSSL matches or surpasses state-of-the-art self-supervised learning methods, highlighting its robustness and generalization capabilities. Beyond performance, VSSL contributes a new perspective to the design of latent variable models by bridging probabilistic modeling and self-supervised learning—without relying on generative reconstruction. This work opens new directions for lightweight, probabilistically grounded SSL methods that are both theoretically sound and practically effective.

\section*{Acknowledgment}

The numerical calculations reported in this paper were fully performed at TUBITAK ULAKBIM, High Performance and Grid Computing Center (TRUBA resources).

\bibliographystyle{IEEEtran}
\bibliography{mybibfile}

\end{document}